\documentclass[onecolumn]{IEEEtran}
\ifCLASSOPTIONcompsoc
\else
\fi
\ifCLASSINFOpdf
\else
\fi
\usepackage{amsthm}
\usepackage{amssymb}
\usepackage{amsbsy}
\usepackage{amsmath}
\usepackage{natbib}
\usepackage{epsf}
\usepackage{amsmath, amsthm, amssymb}
\usepackage{latexsym}
\usepackage{amsbsy}
\usepackage{graphicx}
\usepackage{flafter}
\usepackage{mohak-def}
\usepackage{color}
\usepackage{subfigure}

\newcommand{\KL}{{\rm KL}}
\newcommand{\kl}{{\rm kl}}

\newtheorem{theorem}{Theorem}

\begin{document}
%
\title{Feature Selection with Conjunctions of Decision Stumps and Learning from Microarray Data}
%
%
%
%

\author{Mohak~Shah,~Mario~Marchand,~and~Jacques~Corbeil
\IEEEcompsocitemizethanks{\IEEEcompsocthanksitem M. Shah is with the Centre for Intelligent Machines, McGill University, Montreal,
Canada, H3A 2A7.\protect\\
E-mail: mohak@cim.mcgill.ca
\IEEEcompsocthanksitem M. Marchand is with the Department of Computer Science and Software Engineering, Pav. Adrien Pouliot, Laval University, Quebec, Canada, G1V-0A6.\protect\\
Email: Mario.Marchand@ift.ulaval.ca
\IEEEcompsocthanksitem J. Corbeil is with CHUL Research Center, Laval University, Quebec (QC) Canada, G1V-4G2.\protect\\
Email: Jacques.Corbeil@crchul.ulaval.ca
}
\thanks{}}

\IEEEcompsoctitleabstractindextext{%
\begin{abstract}
One of the objectives of designing feature selection learning algorithms is to obtain classifiers that depend on a small number of attributes \emph{and} have verifiable future performance guarantees. There are few, if any, approaches that successfully address the two goals \emph{simultaneously}. Performance guarantees become crucial for tasks such as microarray data analysis due to very small sample sizes resulting in limited empirical evaluation. To the best of our knowledge, such algorithms that give theoretical bounds on the future performance have not been proposed so far in the context of the classification of gene expression data. In this work, we investigate the premise of learning a conjunction (or disjunction) of \emph{decision stumps} in Occam's Razor, Sample Compression, and PAC-Bayes learning settings for identifying a small subset of attributes that can be used to perform reliable classification tasks. We apply the proposed approaches for gene identification from DNA microarray data and compare our results to those of well known successful approaches proposed for the task. We show that our algorithm not only finds hypotheses with much smaller number of genes while giving competitive classification accuracy but also have tight risk guarantees on future performance unlike other approaches. The proposed approaches are general and extensible in terms of both designing novel algorithms and application to other domains.
\end{abstract}

}

\maketitle

\IEEEdisplaynotcompsoctitleabstractindextext

%
\IEEEpeerreviewmaketitle

\section{Introduction}

An important challenge in the problem of classification of
high-dimensional data is to design a learning algorithm
that can construct an accurate classifier that depends on the
smallest possible number of attributes. Further, it is often desired that there be realizable guarantees associated with the future performance of such feature selection approaches. With the recent explosion in various technologies generating huge amounts of measurements, the problem of obtaining learning algorithms with performance guarantees has acquired a renewed interest.

Consider the case of biological domain where the advent of microarray technologies~\citep{eisen-99,Lipshutz-et-al-99} have revolutionized the outlook on the
investigation and analysis of genetic diseases. In parallel, on the
classification front, many interesting results have appeared aiming
to distinguish between two or more types of cells, (e.g. diseased
vs. normal, or cells with different types of cancers) based on gene
expression data in the case of DNA microarrays (see, for instance,~\citep{a-99} for results on
Colon Cancer,~\citep{g-99} for Leukaemia). Focusing on very
few genes to give insight into the class association for a microarray sample is quite important owing to a variety of reasons.
For instance, a small subset of genes is easier to analyze as
opposed to the set of genes output by the DNA microarray chips. It also makes it relatively easier
to deduce biological relationships among them as well as study their
interactions. An approach able to identify a very few number
of genes can facilitate customization of chips and validation experiments-- making the
utilization of microarray technology cheaper, affordable, and
effective.

In the view of a diseased versus a normal sample, these genes can be
considered as indicators of the disease's cause. Subsequent validation study
focused on these genes, their behavior, and their interactions, can lead to
better understanding of the disease. Some attempts in
this direction have yielded interesting results. See, for instance,
a recent algorithm proposed by~\citet{wang-et-al-07} involving the
identification of a gene subset based on importance ranking and
subsequently combinations of genes for classification. Another
example is the approach of~\citet{Tibshirani-et-al-03} based on
nearest shrunken centroids. Some kernel based approaches such as the BAHSIC algorithm~\citep{song-07} and their extensions (e.g.,~\citep{sc-09} for short time-series domains) have also appeared.

The traditional methods used for classifying high-dimensional data
are often characterized as either ``filters''
(e.g.~\citep{fcdbsh-00,wang-et-al-07} or ``wrappers''
(e.g.~\citep{gwbv-02}) depending on whether the attribute selection
is performed independent of, or in conjunction with, the base learning
algorithm.

Despite the acceptable empirical results achieved by such
approaches, there is no theoretical justification of their
performance nor do they come with a guarantee on how well will they perform in the future. What is really needed is a
learning algorithm that has \emph{provably good performance guarantees} in the presence of many irrelevant
attributes. This is the focus of the work presented here. 

\subsection{Contributions}

The main contributions of this work come in the form of formulation of feature selection strategies within well established learning settings resulting in learning algorithms that combine the tasks of feature selection and discriminative learning. Consequently, we obtain feature selection algorithms for classification with tight realizable guarantees on their generalization error. The proposed approaches are a step towards more general learning strategies that combine feature selection with the classification algorithm \emph{and} have tight realizable guarantees. We apply the approaches to the task of classifying microarray data where the attributes of the data sample correspond to the expression level measurements of various genes. In fact the choice of decision stumps as learning bias has in part motivated by this application. The framework is general and extensible in a variety of ways. For instance, the learning strategies proposed in this work can readily be extended to other similar tasks that can benefit from this learning bias. An immediate example would be classifying data from other microarray technologies such as in the case of Chromatin Immunoprecipitation experiments. Similarly, learning biases other than the conjunctions of decision stumps, can also be explored in the same frameworks leading to novel learning algorithms.

\subsection{Motivation}

For learning the class of conjunctions of features, we draw motivation from the guarantee that exists for this class in the following form: \emph{if there exists a conjunction, that depends on $r$ out of the $n$ input attributes and that correctly classifies a training set of $m$ examples, then the greedy covering algorithm of~\citet{h-88} will find a conjunction of at most $r\ln m$ attributes that makes no training errors}. Note the absence of dependence on the number $n$ of input attributes. The method is guaranteed to find at most $r\ln m$ attributes and, hence, depends on the number of available samples $m$ but not on the number of attributes $n$ to be analyzed.

We propose learning algorithms for building small conjunctions of \emph{decision stumps}. We examine three approaches to obtain an
optimal classifier based on this premise that mainly vary in the coding strategies for the threshold of each decision stump. The
first two approaches attempt to do this by encoding the threshold either with message strings (Occam's Razor) or by using training
examples (Sample Compression). The third strategy (PAC-Bayes) attempts to examine if an optimal classifier can be obtained by
trading off the sparsity\footnote{This refers to the number of decision stumps used.} of the classifier with the magnitude of the separating margin of each decision stump. In each case, we derive an upper bound on the generalization error of the classifier and subsequently use it to guide the respective algorithm. Finally, we present empirical results on the microarray data classification tasks 
and compare our results to the state-of-the-art approaches proposed for the task including the Support Vector Machine (SVM) coupled with feature selectors, and Adaboost. The preliminary results of this work appeared in~\citep{ms-05}.

\subsection{Organization}

Section~\ref{sec-Definitions-Hard-Greedy-Rays} gives the basic definitions and notions of the learning setting that we utilize and also characterizes the hypothesis class of conjunctions of decision stumps. All subsequent learning algorithms are proposed to learn this hypothesis class. Section~\ref{sec:Occam's_Razor_approach} proposes an Occam's Razor approach to learn conjunctions of decision stumps leading to an upper bound on the generalization error in this framework. Section~\ref{sec-sample-compression-bound-Rays} then proposes an alternate encoding strategy for the message strings using the Sample Compression framework and gives a corresponding risk bound. In Section~\ref{sec:PAC-Bayes_approach}, we propose a PAC-Bayes approach to learn conjunction of decision stumps that enables the learning algorithm to perform an explicit non-trivial margin-sparsity trade-off to obtain more general classifiers. Section~\ref{sec:learning_algos} then proposes algorithms to learn in the three learning settings proposed in Sections~\ref{sec:Occam's_Razor_approach},~\ref{sec-sample-compression-bound-Rays} and~\ref{sec:PAC-Bayes_approach} along with a time complexity analysis. Note that the learning (optimization) strategies proposed in Section~\ref{sec:learning_algos} do not affect the  respective theoretical guarantees of the learning settings. The algorithms are evaluated empirically on real world microarray datasets in Section~\ref{sec:empirical-results}. Section~\ref{sec:analysis} presents a discussion on the results and also provides an analysis of the biological relevance of the selected genes in the case of each dataset, and their agreement with published findings. Finally, we conclude in Section~\ref{sec:conclusion}.

\section{Definitions}\label{sec-Definitions-Hard-Greedy-Rays}

The input space $\Xcal$ consists of all $n$-dimensional vectors
$\xb=(x_1,\ldots, x_n)$ where each real-valued component $x_i\in
[A_i, B_i]$ for $i=1,\ldots n$. Each attribute $x_i$ for instance can refer to the
expression level of gene $i$. Hence, $A_i$ and $B_i$ are,
respectively, the {\em a priori} lower and upper bounds on values
for $x_i$. The output space $\Ycal$ is the set of classification
labels that can be assigned to any input vector $\xb\in\Xcal$. We
focus here on binary classification problems. Thus $\Ycal =
\{0,1\}$. Each example $\zb=(\xb,y)$ is an input vector $\xb \in \Xcal$ with
its classification label $y\in\Ycal$ chosen \emph{i.i.d.} from an unknown distribution $D$ on $\Xcal \times \Ycal$.
The true risk $R(f)$ of any classifier $f$ is
defined as the probability that it misclassifies an example drawn
according to~$D$:
\[
R(f) \eqdef \pr_{\ex\sim D}\LP f(\xb)\ne y \RP = \Eb_{\ex\sim D}
I(f(\xb)\ne y)
\]
where $I(a)=1$ if predicate $a$ is true and $0$ otherwise. Given a
training set $S = \{\zb_1,\ldots,\zb_m\}$ of $m$ examples, the {\em
empirical risk\/} $R_S(f)$ on $S$, of any classifier $f$, is defined
according to:
\[
R_S(f) \eqdef \frac{1}{m}\sum_{i=1}^m I(f(\xb_i)\ne
y_i)\eqdef\Eb_{\ex\sim S} I(f(\xb)\ne y)
\]
The goal of any learning algorithm is to find the classifier with minimal true risk based on measuring empirical risk (and other properties) on the training sample $S$.

We focus on learning algorithms that construct a \emph{conjunction
of decision stumps} from a training set. Each {\em decision stump\/}
is just a threshold classifier defined on a single attribute
(component) $x_k$. More formally, a decision stump is identified by
an {\em attribute index\/} $k\in\{1,\ldots,n\}$, a {\em threshold
value\/} $t \in \Reals$, and a {\em direction\/} $d\in\{-1,+1\}$
(that specifies whether class 1 is on the largest or smallest values
of $x_k$). Given any input example $\xb$, the output $r_{td}^k(\xb)$
of a decision stump is defined as:
\[
r_{td}^k(\xb)\eqdef \LC \begin{array}{lll} 1 & \text{if} &
(x_k-t)d > 0\\
0 & \text{if} & (x_k-t)d \le 0
\end{array} \right.
\]

We use a vector $\kb \eqdef (k_1,\ldots, k_\nkb)$ of attribute
indices $k_j\in\{1,\ldots,n\}$ such that $k_1<k_2<\ldots<k_\nkb$
where $\nkb$ is the number of indices present in $\kb$ (and thus the
number of decision stumps in the conjunction)~\footnote{Although it
is possible to use up to two decision stumps on any attribute, we
limit ourselves here to the case where each attribute can be used
for only one decision stump.}. Furthermore, We use a vector $\tb =
(t_{k_1}, t_{k_2},\ldots, t_{k_\nkb})$ of threshold values and a
vector $\db = (d_{k_1}, d_{k_2},\ldots, d_{k_\nkb})$ of directions
where $k_j\in\{1,\ldots,n\}$ for $j\in\{1,\ldots,\nkb\}$. On any
input example $\xb$, the output $C_{\tb\db}^\kb(\xb)$ of a
conjunction of decision stumps is given by:
\[
C_{\tb\db}^\kb(\xb)\eqdef \LC \begin{array}{lll} 1 & \text{if} &
r_{t_jd_j}^j(\xb)=1\quad\forall j\in\kb\\
0 & \text{if} & \exists j\in\kb : r_{t_jd_j}^j(\xb)=0
\end{array} \right.
\]

Finally, any algorithm that builds a conjunction can be used to
build a disjunction just by exchanging the role of the positive and
negative labeled examples. In order to keep our description simple,
we describe here only the case of a conjunction. However, the case
of disjunction follows symmetrically.

\section{An Occam's Razor Approach}\label{sec:Occam's_Razor_approach}

Our first approach towards learning the conjunction (or disjunction)
of \emph{decision stumps} is the Occam's Razor approach. Basically,
we wish to obtain a hypothesis that can be coded using the least
number of bits. We first propose an Occam's Razor risk bound which
will ultimately guide the learning algorithm.

In the case of zero-one loss, we can model the risk of the classifier as a binomial. Let $\bin(\kappa,m,r)$ be the the binomial tail associated
with a classifier of (true) risk $r$. Then $\bin(\kappa,m,r)$ is the probability
that this classifier makes at most $\kappa$ errors on a set of
$m$ examples:
\[
\bin\LP\kappa,m,r\RP\eqdef \sum_{i=0}^\kappa{m\choose i} r^i
(1-r)^{m-i}
\]
The
\emph{binomial tail inversion} $\binb\LP\kappa,m,\dt\RP$ then gives the largest risk value that a classifier can have while still having
a probability of at least $\dt$ of observing at most $\kappa$ errors
out of $m$ examples~\citep{l-05,bl-03}:
\begin{equation*}\label{bti}
\binb\LP\kappa,m,\dt\RP\eqdef\sup\LC r:
\bin\LP\kappa,m,r\RP\ge\dt\RC
\end{equation*}
From this definition, it follows that $\binb\LP mR_S(f),m,\dt\RP$ is
the \emph{smallest} upper bound, which holds with probability at
least $1-\dt$, on the true risk of any classifier $f$ with an
observed empirical risk $R_S(f)$ on a test set of $m$ examples:
\begin{equation*}\label{tsb}
    \forall f\colon\quad\pr_{S\sim D^m}\LP R(f)\le \binb\bigl(mR_{S}(f),m,\dt\bigr) \RP\ge
    1-\dt
\end{equation*}

Our starting point is the Occam's razor bound of~\citet{l-05} which
is a tighter version of the bound proposed by~\citet{behw-87}. It is
also more general in the sense that it applies to any prior
distribution $P$ over any countable class of classifiers.

\begin{theorem}[\citet{l-05}]\label{occam}
For any prior distribution $P$ over any countable class $\Fcal$ of
classifiers, for any data-generating distribution $D$, and for any $\dt\in(0,1]$, we have:
\[
\pr_{S\sim D^m}\biggl\{\forall f\in\Fcal\colon
    R(f)\le\binb\bigl(
    mR_{S}(f),m,P(f)\dt\bigr)\biggr\}
    \ge 1-\dt
\]
\end{theorem}
The proof (available in~\citep{l-05}) directly follows from a
straightforward union bound argument and from the fact that
$\sum_{f\in\Fcal}P(f)=1$. To apply this bound for conjunctions of
decision stumps we thus need to choose a suitable prior $P$ for this
class. Moreover, Theorem~\ref{occam} is valid when $\sum_{f\in\Fcal}P(f) \le 1$. Consequently, we will use a \emph{subprior} $P$ whose sum is $\le 1$.

In our case, decision-stumps' conjunctions are
specified in terms of the discrete-valued vectors $\kb$ and
$\db$ and the continuous-valued vector $\tb$. We will see below that we
will use a finite-precision bit string $\sg$ to specify the set of
threshold values $\tb$. Let us denote by $P(\kb,\db,\sg)$ the prior
probability assigned to the conjunction $C_{\sg\db}^\kb$ described
by $(\kb,\db,\sg)$. We choose a prior of the following form:
\[
P(\kb,\db,\sg) = \frac{1}{{n\choose\nkb}}p(\nkb)\frac{1}{2^\nkb}
g_{\kb,\db}(\sg)
\]
where $g_{\kb,\db}(\sg)$ is the prior probability assigned to string
$\sg$ given that we have chosen $\kb$ and $\db$. Let
$\Mcal(\kb,\db)$ be the set of all message strings that we can use
given that we have chosen $\kb$ and $\db$. If $\Ical$ denotes the
set of all $2^n$ possible attribute index vectors and $\Dcal_\kb$
denotes the set of all $2^\nkb$ binary direction vectors $\db$ of
dimension $\nkb$, we have that $\sum_{\kb\in\Ical}\sum_{\db\in
\Dcal_\kb}\sum_{\sg\in\Mcal(\kb,\db)}P(\kb,\db,\sg) \le 1$ whenever $\sum_{d=0}^n p(d) \le 1$ and $\sum_{\sg\in\Mcal(\kb,\db)}
g_{\kb,\db}(\sg)\le 1\ \forall\kb,\db$.

The reasons motivating this choice for the prior are the following.
The first two factors come from the belief that the final
classifier, constructed from the group of attributes specified by
$\kb$, should depend only on the number $\nkb$ of attributes in this
group. If we have complete ignorance about the number of decision
stumps the final classifier is likely to have, we should choose
$p(d) = 1/(n+1)$ for $d\in\{0,1,\ldots,n\}$. However, we should
choose a $p$ that decreases as we increase $d$ if we have reasons to
believe that the number of decision stumps of the final classifier
will be much smaller than $n$. Since this is usually our case, we
propose to use:
\[
p(\nkb) = \frac{6}{\pi^2}(\nkb + 1)^{-2}
\]
The third factor of $P(\kb,\db,\sg)$ gives equal prior probabilities
for each of the two possible values of direction $d_j$.

To specify the distribution of strings $g_{\kb,\db}(\sg)$, consider
the problem of coding a threshold value $t\in [a,b]$ $\subset [A,B]$
where $[A,B]$ is some predefined interval in which we are permitted
to choose $t$ and where $[a,b]$ is an interval of ``equally good''
threshold values.\footnote{By a ``good'' threshold value, we mean a
threshold value for a decision stump that would cover many negative
examples and very few positive examples (see the learning
algorithm).} We propose the following diadic coding scheme for the
identification of a threshold value that belongs to that interval.
Let $l$ be the number of bits that we use for the code. Then, a code
of $l$ bits specifies one value among the set $\Ld_l$ of threshold
values:
\[
\Ld_l\eqdef\LC\LB 1-\frac{2j-1}{2^{l+1}}\RB
A+\frac{2j-1}{2^{l+1}}B\RC_{j=1}^{2^l}
\]
We denote by $A_i$ and $B_i$, the respective \emph{a priori} minimum and maximum values that the attribute $i$ can take. These values are obtained from the definition of data. Hence, for an attribute $i \in \kb$, given an interval $[a_i,b_i]\subset [A_i,B_i]$ of threshold values, we take the smallest number $l_i$ of bits such that there exists a threshold
value in $\Ld_{l_i}$ that falls in the interval $[a_i,b_i]$. In that way, we
will need at most $\lfloor\log_2((B_i-A_i)/(b_i-a_i)) \rfloor$ bits to
obtain a threshold value that falls in $[a_i,b_i]$.

Hence, to specify the threshold for each decision stump $i \in \kb$, we need to
specify the number $l_i$ of bits and a $l_i$-bit string $s_i$ that
identifies one of the threshold values in $\Ld_{l_i}$. The risk bound does not depend on how we actually code $\sg$ (for some
receiver). It only depends on the a priori probabilities we assign
to each possible realization of $\sg$. We choose the following
distribution:
\begin{eqnarray}
\label{g-sigma}g_{\kb,\db}(\sg) & \eqdef & g_{\kb,\db}(l_1,s_1,\ldots,l_\nkb,s_\nkb)\\
&=&\prod_{i\in\kb}\zeta(l_i)\cdot 2^{-l_i}
\end{eqnarray}
where:
\begin{equation}\label{zeta}\zeta(a)\ \eqdef\
\frac{6}{\pi^2}(a+1)^{-2}\quad\forall a\in\Naturals
\end{equation}
The sum over all the possible realizations of $\sg$ gives 1 since
$\sum_{i=1}^{\infty} i^{-2} = \pi^2/6$. Note that by giving equal a
priori probability to each of the $2^{l_i}$ strings $s_i$ of length
$l_i$, we give no preference to any threshold value in $\Ld_{l_i}$.

The distribution $\zeta$ that we have chosen for each string length
$l_i$ has the advantage of decreasing slowly so that the risk bound
does not deteriorate too rapidly as $l_i$ increases. Other choices
are clearly possible. However, note that the dominant contribution comes from the $2^{-l_i}$ term yielding a risk bound that depends linearly in $l_i$.

With this choice of prior, we have the following theorem:
\begin{theorem}\label{c-occam}
Given all our previous definitions and for any $\dt\in(0,1]$, we
have:
\[
\Pr_{S\sim D^m}\bigg(\forall\kb,\db,\sg\colon R(C^\kb_{\sg\db})\le\binb\biggl(mR_{S}(C^\kb_{\sg\db}),m,\frac{p(\nkb)g_{\kb,\db}(\sg)\dt}{{n\choose\nkb}2^\nkb}\biggr)\bigg)\geq 1-\dt
\]
\end{theorem}

Finally, we emphasize that the risk bound of Theorem~\ref{c-occam},
used in conjunction with the distribution of messages given by
$g_{\kb,\db}(\sg)$, provides a guide for choosing the optimal
classifier. Note that the above risk bound suggests a non-trivial trade-off between the number of attributes and the length of the message string used to encode the classifier. Indeed the risk bound may be smaller for a conjunction having a large number of attributes with small message strings (i.e., small $l_i$s) than for a conjunction having a small number of attributes but with large message strings.

\section{A Sample Compression Approach}\label{sec-sample-compression-bound-Rays}

The basic idea of the Sample compression framework~\citep{kw-07} is
to obtain learning algorithms with the property that the generated
classifier (with respect to some training data) can often be
reconstructed with a very small subset of training examples. More formally,
a learning algorithm $A$ is said to be a {\em sample-compression
algorithm\/} iff there exists a {\em compression function\/} $\Ccal$
and a {\em reconstruction function\/} $\Rcal$ such that for any
training sample $S = \{\zb_1,\ldots,\zb_m\}$ (where $\zb_i\eqdef
(\xb_i,y_i)$), the classifier $A(S)$ returned by $A$ is given by:
\[
A(S) = \Rcal(\Ccal(S))\quad\forall S\in(\Xcal\times\Ycal)^m
\]

For a training set $S$, the compression function $\Ccal$ of learning
algorithm $A$ outputs a subset $\zb_\ib$ of $S$, called the {\em
compression set\/}, and an {\em information message\/}~$\sg$, i.e., $(\zb_\ib,\sg) = \Ccal(\zb_1,\ldots,\zb_m)$.
The information message $\sg$ contains the additional information
needed to reconstruct the classifier from the compression set
$\zb_\ib$. Given a training sample $S$, we define the compression
set $\zb_\ib$ by a vector of indices $\ib$ such that $\ib \eqdef  (i_1,i_2,\ldots,i_\nib)$, with $i_j\in \{1,\ldots,m\} \forall j$ and $i_1 < i_2 <\ldots< i_\nib$ and where $\nib$ denotes the number of indices present in $\ib$.

When given an arbitrary compression set $\zb_\ib$ and an arbitrary
information message $\sg$, the {\em reconstruction function} $\Rcal$
of a learning algorithm $A$ must output a classifier. The
information message $\sg$ is chosen from a set $\Mcal(\zb_\ib)$ that consists of all the distinct messages that
can be attached to the compression set $\zb_\ib$. The existence
of this reconstruction function $\Rcal$ assures that the classifier
returned by $A(S)$ is {\em always\/} identified by a compression
set $\zb_\ib$ and an information message~$\sg$.

In sample compression settings for learning decision stumps'
conjunctions, the message string consists of the attributes and
directions defined above. However, the thresholds
are now specified by training examples. Hence, if we have $|\kb|$
attributes where $\kb$ is the set of thresholds, the compression set
consists of $|\kb|$ training examples (one per threshold).

Our starting point is the following generic Sample Compression bound
~\citep{msok-05}:
\begin{theorem}\label{Theom:sampleCompression}
For any sample compression learning algorithm with a reconstruction
function $\Rcal$ that maps arbitrary subsets of a training set and
information messages to classifiers:
\[
\Pb_{S\sim D^m}\LC\forall\ib\in\Ical, \sg\in\Mcal(\Zb_\ib)\colon R(\Rcal(\sg,\Zb_\ib))\le\ep(\sg,\Zb_\ib,\njb)\RC\ge 1-\dt
\]
where
\begin{eqnarray}\label{eps2-sol-repeat}
    \ep(\sg,\zb_\ib,\njb) &=& 1 - \exp\LP\frac{-1}{m-\nib-\njb}\LB\ln{m \choose \nib} + \ln{m-\nib \choose \njb}\right.\right.\nonumber\\
    &&\left.\left. + \ln\LP\frac{1}{P_{\Mcal(\Zb_\ib)}(\sg)}\RP +\ln\LP\frac{1}{\zeta(\nib)\zeta(\njb)\dt}\RP\RB\RP\nonumber\\
\end{eqnarray}
and $\zeta$ is defined by Equation~\ref{zeta}.
\end{theorem}
Now, we need to specify the distribution of messages
($P_{\Mcal(\Zb_\ib)}(\sg)$) for the conjunction of decision stumps.
Note that in order to specify a conjunction of decision stumps, the
compression set consists of one example per decision stump. For each
decision stump we have one attribute and a corresponding
threshold value determined by the numerical value that this
attribute takes on the training example.

The learner chooses an attribute whose threshold is identified by the associated training example. The set of these training examples form the compression set. Finally, the learner chooses a direction for each attribute.

The subset of attributes that specifies the decision stumps in
our compression set $\zb_\ib$ is given by the vector $\kb$ defined in the previous section. Moreover, since there is one
decision stump corresponding to each example in the compression set,
we have $\nib = \nkb$. Now, we assign equal probability to each
possible set $\nkb$ of attributes (and hence thresholds) that can be
selected from $n$ attributes. Moreover, we assign equal probability
over the direction that each decision stump can have $(+1, -1)$.
Hence, we get the following distribution of messages:

\begin{equation}\label{dist-messages-Rays}
P_{\Mcal(\zb_\ib)}(\sg) = {n \choose{\nkb}}^{-1} \cdot 2^{-\nkb}\hspace{0.2cm}\forall\sigma
\end{equation}
Equation~\ref{dist-messages-Rays} along with the \emph{Sample
Compression Theorem} completes the bound for the conjunction of
decision stumps.

\section{A PAC-Bayes Approach}\label{sec:PAC-Bayes_approach}

The Occam's Razor and Sample Compression, in a sense, aim at obtaining
sparse classifiers with minimum number of stumps. This sparsity is
enforced by selecting the classifiers with minimal encoding of the message strings and the compression set in respective cases.

We now examine if by sacrificing this sparsity in terms of a larger
separating margin around the decision boundary (yielding more
confidence) can lead us to classifiers with smaller generalization error. The learning
algorithm is based on the PAC-Bayes approach~\citep{m-99} that aims
at providing \textbf{P}robably \textbf{A}pproximately
\textbf{C}orrect (PAC) guarantees to ``Bayesian'' learning
algorithms specified in terms of a {\em prior distribution\/} $P$
(before the observation of the data) and a data-dependent,
{\em posterior distribution\/} $Q$ over a space of classifiers.

We formulate a learning algorithm that outputs a stochastic
classifier, called the \emph{Gibbs Classifier} $G_Q$ defined by a data-dependent posterior $Q$. Our
classifier will be partly stochastic in the sense that we will
formulate a posterior over the threshold values utilized by the
decision stumps while still retaining the deterministic nature for
the selected attributes and directions for the decision stumps.

Given an input example $\xb$, the Gibbs classifier first selects a
classifier $h$ according to the posterior distribution $Q$ and then
use $h$ to assign the label $h(\xb)$ to $\xb$. The risk of $G_Q$ is
defined as the expected risk of classifiers drawn according to $Q$:
\[
R(G_Q)\eqdef \Eb_{h\sim Q} R(h) = \Eb_{h\sim Q}\Eb_{(\xb,y)\sim D}
I(h(\xb)\ne y)
\]
Our starting point is the PAC-Bayes theorem~\citep{m-03,l-05,s-02}
that provides a bound on the risk of the Gibbs classifier:

\begin{theorem}\label{thm:pac-bayes}
Given any space $\Hcal$ of classifiers. For any data-independent
prior distribution $P$ over $\Hcal$, we have:
\[
\Pr_{S\sim D^m}\bigg(\forall Q:\kl(R_S(G_Q)\|R(G_Q))\le \frac{\KL(Q\|P) + \ln\frac{m+1}{\dt}}{m}\bigg)\geq 1-\dt
\]
where $\KL(Q\|P)$ is the Kullback-Leibler divergence between
distributions\footnote{Here $Q(h)$ denotes the probability density
function associated with $Q$, evaluated at $h$.} $Q$ and $P$:
\[
\KL(Q\|P) \eqdef \Eb_{h\sim Q}\ln\frac{Q(h)}{P(h)}
\]
and where $\kl(q\|p)$ is the Kullback-Leibler divergence between the
Bernoulli distributions with probabilities of success $q$ and $p$:
\[
\kl(q\|p) \eqdef q \ln\frac{q}{p} +
(1-q)\ln\frac{1-q}{1-p}
\]
\end{theorem}
This bound for the risk of Gibbs classifiers can easily be turned
into a bound for the risk of Bayes classifiers $B_Q$ over the
posterior $Q$. $B_Q$ basically performs a majority vote (under
measure $Q$) of binary classifiers in $\Hcal$. When $B_Q$
misclassifies an example $\xb$, at least half of the binary
classifiers (under measure $Q$) misclassifies $\xb$. It follows that
the error rate of $G_Q$ is at least half of the error rate of $B_Q$.
Hence $R(B_Q)\le 2R(G_Q)$.

In our case, we have seen that decision stump conjunctions are
specified in terms of a mixture of discrete parameters $\kb$ and
$\db$ and continuous parameters $\tb$. If we denote by
$P_{\kb,\db}(\tb)$ the probability density function associated with
a prior $P$ over the class of decision stump conjunctions, we
consider here priors of the form:
\[
P_{\kb,\db}(\tb) =
\frac{1}{{n\choose\nkb}}p(\nkb)\frac{1}{2^\nkb}\prod_{j\in\kb}\frac{I(t_j \in [A_j,B_j])}{B_j-A_j}
\]

As before, we have that:
\[
\sum_{\kb\in\Ical}\sum_{\db\in
\Dcal_\kb}\prod_{j\in\kb}\int_{A_j}^{B_j}dt_j P_{\kb,\db}(\tb) = 1
\]
whenever $\sum_{e=0}^n p(e) =1$.\\

The factors relating to the discrete components $\kb$ and $\db$ have
the same rationale as in the case of the Occam's Razor approach. However, in the case of
the threshold for each decision stumps, we now consider an
explicitly continuous uniform prior. As in the Occam's Razor case, we assume each attribute value
$x_k$ to be constrained, a priori, in $[A_k,B_k]$ such that $A_k$ and $B_k$ are obtained from the definition of the data. Hence, we have chosen
a uniform prior probability density on $[A_k,B_k]$ for each $t_k$ such
that $k\in\kb$. This explains the last factors of
$P_{\kb,\db}(\tb)$.

Given a training set $S$, the learner will choose an attribute group
$\kb$ and a direction vector $\db$ deterministically. We pose the
problem of choosing the threshold in a similar manner as in the case
of Occam's Razor approach of
Section~\ref{sec:Occam's_Razor_approach} with the only difference
that the learner identifies the interval and selects a threshold
stochastically. For each attribute $x_k\in [A_k, B_k] : k\in\kb$, a
margin interval $[a_k,b_k]\subseteq [A_k, B_k]$ is chosen by the
learner. A deterministic decision stump conjunction classifier is
then specified by choosing the thresholds values $t_k\in [a_k,b_k]$
uniformly. It is tempting at this point to choose $t_k =
(a_k+b_k)/2\ \forall k\in\kb$ (\ie, in the middle of each interval).
However, the PAC-Bayes theorem offers a better guarantee for another
type of deterministic classifier as we see below.

Hence, the Gibbs classifier is defined with a posterior distribution
$Q$ having all its weight on the same $\kb$ and $\db$ as chosen by
the learner but where each $t_k$ is uniformly chosen in $[a_k,b_k]$.
The KL divergence between this posterior $Q$ and the prior $P$ is
then given by:
\[
KL(Q\|P) = \ln\LP{n\choose\nkb}\cdot\frac{2^\nkb}{p(\nkb)}\RP +
\sum_{k\in\kb}\ln\LP\frac{B_k-A_k}{b_k-a_k}\RP
\]

In this limit when $[a_k,b_k]=[A_k,B_k]\ \forall k\in\kb$, it can be
seen that the KL divergence between the ``continuous components'' of
$Q$ and $P$ vanishes. Furthermore, the KL divergence between the
``discrete components'' of $Q$ and $P$ is small for small values of
$\nkb$ (whenever $p(\nkb)$ is not too small). {\em Hence, this KL
divergence between our choices for $Q$ and $P$ exhibits a tradeoff
between margins ($b_k-a_k$) and sparsity (small value of $\nkb$) for
Gibbs classifiers\/}. Theorem~\ref{thm:pac-bayes} suggests that the
$G_Q$ with the smallest guarantee of risk $R(G_Q)$ should minimize a
non trivial combination of $KL(Q\|P)$ and $R_S(G_Q)$.

The posterior $Q$ is identified by an attribute group vector $\kb$,
a direction vector $\db$, and intervals $[a_k,b_k]\ \forall
k\in\kb$. We refine the notation for our Gibbs classifier $G_Q$ to
reflect this. Hence, we use $G_{\ab\bb}^{\kb\db}$ where $\ab$ and
$\bb$ are the vectors formed by the unions of $a_k$s and $b_k$s
respectively. We can obtain a closed-form expression for
$R_S(G_{\ab\bb}^{\kb\db})$ by first considering the risk
$R_\ex(G_{\ab\bb}^{\kb\db})$ on a single example $\ex$ since
$R_S(G_{\ab\bb}^{\kb\db}) = \Eb_{\ex\sim
S}R_\ex(G_{\ab\bb}^{\kb\db})$. From our definition for $Q$, we find
that:

\begin{equation}\label{eq:RG-ex}
R_\ex(G_{\ab\bb}^{\kb\db}) =
(1-2y)\LB\prod_{k\in\kb}\sg_{a_k,b_k}^{d_k}(x_k) - y\RB
\end{equation}
where:
\begin{eqnarray*}
\sg_{a,b}^d(x)\eqdef\LC\begin{array}{ll}
    0 & \text{if}\hspace{0.1cm}(x<a\hspace{0.1cm}\text{and}\hspace{0.1cm}d=+1)\hspace{0.1cm}\text{or}\hspace{0.1cm}(b<x\hspace{0.1cm}\text{and}\hspace{0.1cm}d=-1)\\
    \frac{x-a}{b-a} & \text{if}\hspace{0.1cm}a\le x\le b\hspace{0.1cm}\text{and}\hspace{0.1cm}d=+1\\
    \frac{b-x}{b-a} & \text{if}\hspace{0.1cm}a\le x\le b\hspace{0.1cm}\text{and}\hspace{0.1cm}d=-1\\
    1 & \text{if}\hspace{0.1cm} (b<x\hspace{0.1cm}\text{and}\hspace{0.1cm}d=+1)\hspace{0.1cm}\text{or}\hspace{0.1cm}(x<a\hspace{0.1cm}\text{and}\hspace{0.1cm}d=-1)
    \end{array}\right.
\end{eqnarray*}

Note that the expression for $R_\ex(C_{\tb\db}^\kb)$ is identical to
the expression for $R_\ex(G_{\ab\bb}^{\kb\db})$ except that the
piece-wise linear functions $\sg_{a_k,b_k}^{d_k}(x_k)$ are replaced
by the indicator functions $I((x_k-t_k)d_k>0)$.

The PAC-Bayes theorem provides a risk bound for the Gibbs classifier
$G_{\ab\bb}^{\kb\db}$. Since the Bayes classifier
$B_{\ab\bb}^{\kb\db}$ just performs a majority vote under the same
posterior distribution as the one used by $G_{\ab\bb}^{\kb\db}$, it
follows that:
\begin{equation}\label{eq:B}
B_{\ab\bb}^{\kb\db}(\xb) = \LC\begin{array}{lll}
    1 &\text{if}& \prod_{k\in\kb}\sg_{a_k,b_k}^{d_k}(x_k) > 1/2\\
    0 &\text{if}& \prod_{k\in\kb}\sg_{a_k,b_k}^{d_k}(x_k) \le 1/2
\end{array}\right.
\end{equation}
\newline
\indent Note that $B_{\ab\bb}^{\kb\db}$ has an \emph{hyperbolic}
decision surface. Consequently, $B_{\ab\bb}^{\kb\db}$ is not
representable as a conjunction of decision stumps. There is,
however, no computational difficulty at obtaining the output of
$B_{\ab\bb}^{\kb\db}(\xb)$ for any $\xb\in\Xcal$. We now state our
main theorem:

\begin{theorem}\label{thm:main-Rays}
Given all our previous definitions, for any $\dt\in(0,1]$, and for
any $p$ satisfying $\sum_{e=0}^n p(e) = 1$, we have, with
probability atleast $1-\dt$ over random draws of $S\sim D^m$:
\[
\Big(\forall\kb,\db,\ab,\bb  \colon R(G_{\ab\bb}^{\kb\db}) \le
\sup\LC \ep\colon \kl(R_S(G_{\ab\bb}^{\kb\db})\|\ep ) \le \psi \RC
\Big)
\]
where
\[
\psi =
\frac{1}{m}\LB\ln\LP{n\choose\nkb}\cdot\frac{2^\nkb}{p(\nkb)}\cdot\frac{m+1}{\dt}\RP+\sum_{k\in\kb}\ln\left(\frac{B_k-A_k}{b_k-a_k}\right)\RB\\
\]
Furthermore: $R(B_{\ab\bb}^{\kb\db})\le
2R(G_{\ab\bb}^{\kb\db})\quad\forall\kb,\db,\ab,\bb$.
\end{theorem}
\section{The Learning Algorithms}\label{sec:learning_algos}

Having proposed the theoretical frameworks attempting to obtain the
optimal classifiers based on various optimization criteria, we now
detail the learning algorithms for these approaches. Ideally, we
would like to find a conjunction of decision stumps that minimizes
the respective risk bounds for each approach. Unfortunately, this
cannot be done efficiently in all cases since this problem is at
least as hard as the (NP-hard) minimum set cover
problem as mentioned by~\citet{ms-02}. Hence, we use a set covering greedy
heuristic. It consists
of choosing the decision stump $i$ with the largest {\em utility\/}
$U_i^{SC}$ where:
\begin{equation}\label{hard-utlty-scm-repeat}
 U_i^{SC} = |Q_i| - p|R_i|
\end{equation}
where $Q_i$ is the set of negative examples covered (classified as
0) by feature $i$, $R_i$ is the set of positive examples
misclassified by this feature, and $p$ is a learning parameter that
gives a penalty $p$ for each misclassified positive example. Once
the feature with the largest $U_i$ is found, we remove $Q_i$ and
$R_i$ from the training set $S$ and then repeat (on the remaining
examples) until either no more negative examples are present or that
a maximum number $s$ of features has been reached. This heuristic
was also used by~\citet{ms-02} in the context of a sample
compression classifier called the set covering machine. For our
sample compression approach (SC), we use the above utility function
$U_i^{SC}$.

However, for the Occam's Razor and the PAC-Bayes approaches, we need
utility functions that can incorporate the optimization aspects
suggested by these approaches. 
\subsection{The Occam's Razor learning algorithm}

We propose the following learning strategy for Occam's Razor
learning of conjunctions of decision stumps. For a fixed $l_i$ and
$\eta$, let $N$ be the set of negative examples and $P$ be the set
of positive examples. We start with $N'=N$ and $P'=P$. Let $Q_i$ be
the subset of $N'$ covered by decision stump $i$, let $R_i$ be the
subset of $P'$ covered by decision stump $i$, and let $l_i$ be the
number of bits used to code the threshold of decision stump $i$. We
choose the decision stump $i$ that maximizes the \emph{utility}
$U_i^{Occam}$ defined as:
\begin{eqnarray*}\label{util}
U_i^{Occam} &\eqdef& \frac{|Q_i|}{N'} - p \frac{|R_i|}{P} -
\eta\cdot l_i
\end{eqnarray*}
where $p$ is the \emph{penalty} suffered by covering (and hence,
misclassifying) a positive example and $\eta$ is the cost of using
$l_i$ bits for decision stump $i$. Once we have found a decision
stump maximizing $U_i$, we update $N'=N'-Q_i$ and $P'=P'-R_i$ and
repeat to find the next decision stump until either $N'=\emptyset$
or the maximum number $v$ of decision stumps has been reached (early
stopping the greedy). The best values for the learning parameters $p,
\eta$, and $v$ are determined by cross-validation.

\subsection{The PAC-Bayes Learning Algorithm}

Theorem~\ref{thm:main-Rays} suggests that the learner should try to find the Bayes
classifier $B_{\ab\bb}^{\kb\db}$ that uses a small number of
attributes (\ie, a small $\nkb$), each with a large separating
margin $(b_k-a_k)$, while keeping the empirical Gibbs risk
$R_S(G_{\ab\bb}^{\kb\db})$ at a low value. As discussed earlier, we utilize the greedy set covering heuristic for learning.

In our case, however, we need to keep the Gibbs risk on $S$ low
instead of the risk of a deterministic classifier. Since the Gibbs
risk is a ``soft measure'' that uses the piece-wise linear functions
$\sg_{a,b}^d$ instead of the ``hard'' indicator functions, we cannot
make use of the hard utility function of Equation~\ref{hard-utlty-scm-repeat}.
Instead, we need a ``softer'' version of this utility function to
take into account covering (and erring on) an example partly. That
is, a negative example that falls in the linear region of a
$\sg_{a,b}^d$ is in fact partly covered and vice versa for the
positive example.

Following this observation, let $\kb'$ be the vector of indices of
the attributes that we have used so far for the construction of the
classifier. Let us first define the \emph{covering value}
$\Ccal(G_{\ab\bb}^{\kb'\db})$ of $G_{\ab\bb}^{\kb'\db}$ by the
``amount'' of negative examples assigned to class $0$ by
$G_{\ab\bb}^{\kb'\db}$:
\begin{eqnarray*}
\Ccal(G_{\ab\bb}^{\kb'\db}) &\eqdef& \sum_{\ex\in S} (1-y)\LB 1
-\prod_{j\in\kb'}\sg_{a_j,b_j}^{d_j}(x_j)\RB
\end{eqnarray*}
\newline
We also define the \emph{positive-side error}
$\Ecal(G_{\ab\bb}^{\kb'\db})$ of $G_{\ab\bb}^{\kb'\db}$ as the
``amount'' of positive examples assigned to class $0$ :
\begin{eqnarray*}
\Ecal(G_{\ab\bb}^{\kb'\db}) &\eqdef& \sum_{\ex\in S}  y\LB 1 -
\prod_{j\in\kb'}\sg_{a_j,b_j}^{d_j}(x_j)\RB
\end{eqnarray*}

We now want to add another decision stump on another attribute, call
it $i$, to obtain a new vector $\kb''$ containing this new attribute
in addition to those present in $\kb'$. Hence, we now introduce the
\emph{covering contribution} of decision stump $i$ as:
\begin{eqnarray*}
\Ccal_{\ab\bb}^{\kb'\db}(i) &\eqdef& \Ccal(G_{\ab'\bb'}^{\kb''\db'})
-\Ccal(G_{\ab\bb}^{\kb'\db})\\
    &=& \sum_{\ex\in S}(1-y)\LB 1 - \sg_{a_i,b_i}^{d_i}(x_i) \RB
    \prod_{j\in\kb'}\sg_{a_j,b_j}^{d_j}(x_j)\\
\end{eqnarray*}
and the \emph{positive-side error contribution} of decision stump
$i$ as:
\begin{eqnarray*}
\Ecal_{\ab\bb}^{\kb'\db}(i) &\eqdef& \Ecal(G_{\ab'\bb'}^{\kb''\db'})
-\Ecal(G_{\ab\bb}^{\kb'\db})\\
    &=& \sum_{\ex\in S} y\LB 1 - \sg_{a_i,b_i}^{d_i}(x_i) \RB
\prod_{j\in\kb'}\sg_{a_j,b_j}^{d_j}(x_j)
\end{eqnarray*}

Typically, the covering contribution of decision stump $i$ should
increase its ``utility'' and its positive-side error should decrease
it. Moreover, we want to decrease the ``utility'' of decision stump
$i$ by an amount which would become large whenever it has a small
separating margin. Our expression for $KL(Q\|P)$ suggests that this
amount should be proportional to $\ln((B_i-A_i)/(b_i-a_i))$.
Furthermore we should compare this margin term with the
\emph{fraction} of the remaining negative examples that decision
stump $i$ has covered (instead of the absolute amount of negative
examples covered). Hence the covering contribution
$\Ccal_{\ab\bb}^{\kb'\db}(i)$ of decision stump $i$ should be
divided by the amount $\Ncal_{\ab\bb}^{\kb'\db}$ of negative
examples that \emph{remains} to be covered before considering
decision stump $i$:
\[
\Ncal_{\ab\bb}^{\kb'\db}\eqdef \sum_{\ex\in S}
(1-y)\prod_{j\in\kb'}\sg_{a_j,b_j}^{d_j}(x_j)
\]
which is simply the amount of negative examples that have been
assigned to class~1 by $G_{\ab\bb}^{\kb'\db}$. If $P$ denotes the
set of positive examples, we define the {\em utility
$U_{\ab\bb}^{\kb'\db}(i)$ of adding decision stump $i$ to
$G_{\ab\bb}^{\kb'\db}$\/} as:
\begin{eqnarray*}
   U_{\ab\bb}^{\kb'\db}(i) &\eqdef& \frac{\Ccal_{\ab\bb}^{\kb'\db}(i)}{\Ncal_{\ab\bb}^{\kb'\db}}
   - p \frac{\Ecal_{\ab\bb}^{\kb'\db}(i)}{|P|} - \eta \ln\frac{B_i-A_i}{b_i-a_i}
\end{eqnarray*}
where parameter $p$ represents the \emph{penalty} of misclassifying
a positive example and $\eta$ is another parameter that controls the
importance of having a large margin. These learning parameters can
be chosen by cross-validation. For fixed values of these parameters,
the ``soft greedy'' algorithm simply consists of adding, to the
current Gibbs classifier, a decision stump with maximum added
utility until either the maximum number $v$ of decision stumps has
been reached or all the negative examples have been (totally)
covered. It is understood that, during this soft greedy algorithm,
we can remove an example $\ex$ from $S$ whenever it is totally
covered. This occurs whenever
$\prod_{j\in\kb'}\sg_{a_j,b_j}^{d_j}(x_j)=0$.

Hence, we use the above utility function for the PAC-Bayes learning
strategy. Note that, in the case of $U_i^{PB}$ and $U_i^{Occam}$, we normalize the number of covered and erred examples so as to increase their sensitivity to the respective $\eta$ terms.

\subsubsection{Time Complexity Analysis}

Let us analyze the time complexity of this algorithm for fixed $p$
and $\eta$. For each attribute, we first sort the $m$ examples with respect to their
values for the attribute under consideration. This takes $O(m \log
m)$ time. Then, we examine each potential $a_i$ value (defined by the values of that attribute on the examples). Corresponding
to each $a_i$, we examine all the potential $b_i$ values (all the
values greater than $a_i$). This gives us a time complexity of
$O(m^2)$. Now if $k$ is the largest number of examples falling into
$[a_i,b_i]$, calculating the covering and error contributions and
then finding the best interval $[a_i,b_i]$ takes $O(km^2)$ time.
Moreover, we allow $k \in O(m)$ giving us a time complexity of
$O(m^3)$ for each attribute. Finally, we do this over all the attributes. Hence, the
overall time complexity of the algorithm is $O(nm^3)$. Note, however, that for microarray data, we have $n >> m$ (hence, we can consider $m^3$ to be a constant). Moreover once the best stump is found, we remove the examples covered by this
stump from the training set and repeat the algorithm. Now, we know
that greedy algorithms of this kind have the following guarantee: if
there exist $r$ decision stumps that covers all the $m$ examples,
the greedy algorithm will find at most $r\ln(m)$ decision stumps.
Since we almost always have $r \in O(1)$, the running time of the
whole algorithm will almost always be $\in O(nm^3 \log(m))$. The
good news is, since $n >> m$, the time complexity of our algorithm is roughly linear in $n$.

\subsubsection{Fixed-Margin
Heuristic}\label{subsec:PAC-Bayes-fixed-margin-heuristic}

In order to show why we prefer a uniformly distributed threshold as
opposed to the one fixed at the middle of the interval $\LB a_i,
b_i\RB$ for each stump $i$, we use an alternate algorithm that we call the fixed margin heuristic. The algorithm is similar to
the one described above but with an additional parameter $\gamma$.
This parameter decides a fixed margin boundary around the threshold,
i.e. $\gamma$ decides the length of the interval $\LB a_i, b_i\RB$.
The algorithm still chooses the attribute vector $\kb$, the
direction vector $\db$ and the vectors $\ab$ and $\bb$. However, the
$a_i$'s and $b_i$'s for each stump $i$ are chosen such that, $
|b_i - a_i| = 2\gamma$. The threshold $t_i$ is then fixed in the
middle of this interval, that is $t_i = \frac{(a_i + b_i)}{2}$.
Hence, for each stump $i$, the interval $\LB a_i, b_i\RB = \LB t_i
- \gamma, t_i + \gamma\RB$. For fixed $p$ and $\gamma$, a similar analysis as in the previous subsection yields a time complexity of $O(nm^2\log (m))$ for this algorithm.

\section{Empirical Results}\label{sec:empirical-results}

\begin{table}[t]
\begin{center}
\resizebox{11cm}{!}{
\begin{tabular}{|c|c|c||c||c|c||c|c||c|c|}
\hline \multicolumn{3}{|c||}{\textbf{Data Set}} & \textbf{SVM}
        & \multicolumn{2}{c||}{\textbf{SVM+gs}} & \multicolumn{2}{c||}{\textbf{SVM+rfe}}& \multicolumn{2}{c|}{\textbf{Adaboost}}\\
\hline \textbf{Name} & \textbf{{\small ex}} & \textbf{{\small Genes}}& \textbf{{\small
Errs}} & \textbf{{\small Errs}} & \textbf{S} &\textbf{{\small Errs}} &\textbf{S}& {\textbf{Itrs}} & {\textbf{Errs}}\\
\hline
{\small Colon}    & 62 & 2000 & 12.8$\pm$1.4 & {\small 14.4$\pm$3.5} &  256 &15.4$\pm$4.8&128  &  20 & 15.2$\pm$2.1\\
{\small B\_MD}    & 34 & 7129 & 13.2$\pm$1 &  {\small 7.2$\pm$2.6} &   32 &10.4$\pm$2.4&64   &  20 & 9.8$\pm$1.1\\
{\small C\_MD}    & 60 & 7129 & 28.2$\pm$2.2 & {\small 23.1$\pm$2.8} & 1024 &28.2$\pm$2.2&7129  & 50 & 21.2$\pm$2.4\\
{\small Leuk}  & 72 & 7129 & 21.3$\pm$1.4 & {\small 14$\pm$2.8} &   64&21$\pm$3.2&256   & 20 & 17.8$\pm$1.8\\
{\small Lung}     & 52 & 918 & 8.8$\pm$1.3 & {\small 6.8$\pm$1.9} & 64&7.2$\pm$1.8&32        & 1 & 2.4$\pm$1.4\\
{\small BreastER} & 49 & 7129 & 15.3$\pm$2.4 & {\small 10.3$\pm$2.7} & 256&11.2$\pm$2.8&256      & 50 & 9.8$\pm$1.7\\
\hline
\end{tabular}
}
\caption{Results of SVM, SVM coupled with Golub's feature selection algorithm (filter), SVM with Recursive Feature Elimination (wrapper) and Adaboost algorithms on Gene
Expression datasets.} \label{tab:results-benchmarks}
\end{center}
\end{table}
\begin{table}[t]
\begin{center}
\resizebox{10cm}{!}{
\begin{tabular}{|c|c|c||c|c|c||c|c|}
\hline \multicolumn{3}{|c||}{\textbf{Data Set}} & \multicolumn{3}{c||}{\textbf{Occam}}&\multicolumn{2}{c|}{\textbf{SC}}\\
\hline \textbf{Name} & \textbf{{\small ex}} & \textbf{{\small Genes}} & \textbf{Errs} & \textbf{S} & \textbf{bits} & \textbf{Errs} & \textbf{S}\\
\hline
{\small Colon}    & 62 & 2000 & 23.6$\pm$1.2 & 1.8$\pm$.6 & 6 &18.2$\pm$1.8 &1.2$\pm$.6\\
{\small B\_MD}    & 34 & 7129 &  17.2$\pm$1.8 & 1.2$\pm$.8 & 3&17.2$\pm$1.3&1.4$\pm$.8\\
{\small C\_MD}    & 60 & 7129 & 28.6$\pm$1.8 & 2.6$\pm$1.1 & 4&29.2$\pm$1.1&1.2$\pm$.6\\
{\small Leuk}  & 72 & 7129 & 27.8$\pm$1.7 & 2.2$\pm$.8 & 6&27.3$\pm$1.7&1.4$\pm$.7\\
{\small Lung}  & 52 & 918 & 21.7$\pm$1.1 & 1.8$\pm$1.2 & 5&18$\pm$1.3&1.2$\pm$.5\\
{\small BreastER} & 49 & 7129 & 25.4$\pm$1.2 & 3.2$\pm$.6 & 2&21.2$\pm$1.5&1.4$\pm$.5\\
\hline
\end{tabular}
}
\caption{Results of the proposed Occam's Razor and Sample Compression learning algorithms on Gene
Expression datasets.} \label{tab:results-new-approaches-1}
\end{center}
\end{table}
\begin{table}[t]
\begin{center}
\resizebox{9cm}{!}{
\begin{tabular}{|c|c|c||c|c|c|}
\hline \multicolumn{3}{|c||}{\textbf{Data Set}} & \multicolumn{3}{c|}{\textbf{PAC-Bayes}}\\
\hline \textbf{Name} & \textbf{{\small ex}} & \textbf{{\small Genes}} & {\textbf{S}} & {\small{\textbf{G-errs}}} & {\small{\textbf{B-errs}}} \\
\hline
{\small Colon}    & 62 & 2000 & 1.53$\pm$.28 & 14.68$\pm$1.8 &  14.65$\pm$1.8\\
{\small B\_MD}    & 34 & 7129 & 1.2$\pm$.25 &  8.89$\pm$1.65 &  8.6$\pm$1.4\\
{\small C\_MD}    & 60 & 7129 & 3.4$\pm$1.8 & 23.8$\pm$1.7 & 22.9$\pm$1.65\\
{\small Leuk}  & 72 & 7129 & 3.2$\pm$1.4 & 24.4$\pm$1.5 & 23.6$\pm$1.6\\
{\small Lung}  & 52 & 918 & 1.2$\pm$.3 & 4.4$\pm$.6 & 4.2$\pm$.8\\
{\small BreastER} & 49 & 7129 & 2.6$\pm$1.1 & 12.8$\pm$.8 & 12.4$\pm$.78\\
\hline
\end{tabular}
}
\caption{Results of the PAC-Bayes learning algorithm on Gene
Expression datasets.} \label{tab:results-new-approaches-2}
\end{center}
\end{table}

The proposed approaches for learning conjunctions of \emph{decision
stumps} were tested on the six real-world binary microarray datasets viz. the \emph{colon tumor}~\citep{a-99}, the
\emph{Leukaemia}~\citep{g-99}, the \emph{B}\_\emph{MD} and
\emph{C}\_\emph{MD} Medulloblastomas data~\citep{p-02}, the
\emph{Lung}~\citep{g-01}, and the \emph{BreastER} data~\citep{w-01}.

The \emph{colon tumor} data set~\citep{a-99} provides the expression
levels of 40 tumor and 22 normal colon tissues measured for 6500
human genes. We use the set of 2000 genes identified to have the
highest minimal intensity across the 62 tissues. The \emph{Leuk} data set~\citep{g-99} provides the expression levels
of 7129 human genes for 47 samples of patients with Acute
Lymphoblastic Leukemia (ALL) and 25 samples of patients with Acute
Myeloid Leukemia (AML). The \emph{B}\_\emph{MD} and \emph{C}\_\emph{MD} data
sets~\citep{p-02} are microarray samples containing the expression
levels of 7129 human genes. Data set \emph{B}\_\emph{MD} contains 25
classic and 9 desmoplastic medulloblastomas whereas data set
\emph{C}\_\emph{MD} contains 39 medulloblastomas survivors and 21
treatment failures (non-survivors). The \emph{Lung} dataset consists of gene expression levels of 918
genes of 52 patients with 39 Adenocarcinoma and 13 Squamous Cell
Cancer~\citep{g-01}. This data has some missing values which were
replaced by zeros. Finally, the \emph{BreastER} dataset is the Breast Tumor data of
~\citet{w-01} used with Estrogen Receptor status to label the
various samples. The data consists of expression levels of 7129
genes of 49 patients with 25 positive Estrogen Receptor samples and
24 negative Estrogen Receptor samples.

The number of examples and the number of genes in each data are
given in the ``ex'' and ``Genes'' columns respectively under the ``Data Set'' tab in each table. The algorithms are referred to as
``Occam'' (Occam's Razor), ``SC'' (Sample Compression)
and ``PAC-Bayes'' (PAC-Bayes) in Tables~\ref{tab:results-new-approaches-1} to~\ref{tab:results-PAC-Bayes-bound}.
They utilize the respective theoretical frameworks discussed in
Sections~\ref{sec:Occam's_Razor_approach},~\ref{sec-sample-compression-bound-Rays} and~\ref{sec:PAC-Bayes_approach} along with the respective learning
strategies of Section~\ref{sec:learning_algos}.

We have compared our learning algorithm with a linear-kernel
soft-margin SVM trained both on all the attributes (gene
expressions) and on a subset of attributes chosen by the filter
method of~\citet{g-99}. The filter method consists of ranking the
attributes as function of the difference between the
positive-example mean and the negative-example mean and then use
only the first $\ell$ attributes. The resulting learning algorithm,
named \emph{SVM+gs} is the one used by~\citet{fcdbsh-00} for the
same task. \citet{gwbv-02} claimed obtaining better results with the
recursive feature elimination method but, as pointed out
by~\citet{a-02}, their work contained a methodological flaw. We use
the SVM recursive feature elimination algorithm with this bias
removed and present these results as well for comparison (referred
to as ``SVM+rfe'' in Table~\ref{tab:results-benchmarks}). Finally, we also
compare our results with the state-of-the-art Adaboost algorithm.
For this, we use the implementation in the Weka data mining
software~\citep{wf-05}.

Each algorithm was tested over 20 random permutations of the datasets, with the 5-fold cross validation (CV)
method. Each of the five training sets and testing sets was the same
for all algorithms. The learning parameters of all algorithms and
the gene subsets (for ``SVM+gs'' and ``SVM+rfe'') were chosen from
the training sets \emph{only}. This was done by performing a second
(nested) 5-fold CV on each training set.

For the gene subset selection procedure of SVM+gs, we have
considered the first $\ell = 2^i$ genes (for $i=0,1,\ldots, 12$)
ranked according to the criterion of~\citet{g-99} and have chosen
the $i$ value that gave the smallest 5-fold CV error on the training
set. The ``Errs'' column under each algorithm in Tables~\ref{tab:results-benchmarks} to~\ref{tab:results-new-approaches-2} refer to the average (nested) 5-fold cross-validation error of the
respective algorithm with one standard deviation two-sided confidence interval. The ``bits'' column in Table~\ref{tab:results-new-approaches-1} refer to the number of bits used for the Occam' Razor approach. The ``G-errs'' and the ``B-errs'' columns in Table~\ref{tab:results-new-approaches-2} refer to the average nested 5-fold CV
error of the optimal Gibbs classifier and the corresponding Bayes
classifier with one standard deviation two-sided interval respectively.


For Adaboost, $10$, $20$, $50$, $100$, $200$, $500$, $1000$ and
$2000$ iterations for each datasets were tried and the reported
results correspond to the best obtained 5-fold CV error. 
The size values reported here (the ``S'' columns for ``SVM+gs''and ``SVM+rfe'', and ``Itr'' column for ``AdaBoost'' in Table~\ref{tab:results-benchmarks}) correspond to the number of attributes (genes) selected most frequently by the respective algorithms over all the permutation runs.\footnote{There were no close ties with classifiers with fewer genes.}
Choosing, by cross-validation, the number of boosting
iteration is somewhat inconsistent with Adaboost's goal of
minimizing the empirical exponential risk. Indeed, to comply with
Adaboost's goal, we should choose a large-enough number of boosting
rounds that assures the convergence of the empirical exponential risk
to its minimum value. However, as shown by~\citet{zhang-yu-05}, Boosting is known to overfit when the number of attributes exceeds the number of examples. This happens in the case of microarray experiments frequently where the number of genes far exceeds the number of samples, and is also the case in the datasets mentioned above. Early stopping is the recommended approach in such cases and hence we have followed the method described above to obtained the best number of boosting iterations.

Further, Table~\ref{tab:results-PAC-Bayes-fixed-margin} gives the result for a single run of the deterministic algorithm using the fixed-margin heuristic described above. Table~\ref{tab:results-PAC-Bayes-bound} gives the results for
the PAC-Bayes bound values for the results obtained for a single run of the PAC-Bayes algorithm on the
respective microarray data sets. Recall that the PAC-Bayes bound provides a uniform upper bound on the risk of the Gibbs classifier. The column labels refer to the same quantities as above although the errors reported are over a single nested 5-fold CV run. The ``Ratio'' column of Table~\ref{tab:results-PAC-Bayes-bound} refers to the average value of $(b_k-a_k)/(B_k-A_k)$ obtained over the decision stumps used by the classifiers over $5$ testing folds and the ``Bound'' columns of Tables~\ref{tab:results-PAC-Bayes-fixed-margin}
and~\ref{tab:results-PAC-Bayes-bound} refer to the average risk bound of Theorem~\ref{thm:main-Rays} multiplied by the total number of examples in respective data sets. Note, again, that these results are on a single permutation of the datasets and are presented just to illustrate the practicality of the risk bound and the rationale of not choosing the fixed-margin heuristic over the current learning strategy.

\begin{table}
\begin{center}
\resizebox{10cm}{!}{
\begin{tabular}{|c|c|c||c|c|c|}
\hline \multicolumn{3}{|c||}{\textbf{Data Set}} &  \multicolumn{3}{c|}{\textbf{\small{Stumps:PAC-Bayes(fixed margin)}}}\\
\hline {\textbf{Name}} & {\small{\textbf{ex}}} & {\small{\textbf{Genes}}} & {\textbf{Size}} & {\small{\textbf{Errors}}} & {\textbf{Bound}} \\
\hline
{\small Colon}    & 62 & 2000 &  1 & 14 &   34\\
{\small B\_MD}    & 34 & 7129 & 1 &  7 &  20\\
{\small C\_MD}    & 60 & 7129 &  3 & 28 & 48\\
{\small Leuk}   & 72 & 7129 &  2 & 21 & 46\\
{\small Lung}     & 52 & 918 &  2 & 9 & 29\\
{\small BreastER} & 49 & 7129 & 3 & 11 & 31\\
\hline
\end{tabular}
}
\caption{Results of the PAC-Bayes Approach with
Fixed-Margin Heuristic on Gene Expression Datasets.}
\label{tab:results-PAC-Bayes-fixed-margin}
\end{center}
\end{table}

\begin{table}
\begin{center}
\resizebox{10cm}{!}{
\begin{tabular}{|c|c|c||c||c|c||c|c|c|c|c|}
\hline \multicolumn{3}{|c||}{\textbf{Data Set}} &  \multicolumn{5}{c|}{\textbf{Stumps:PAC-Bayes}}\\
\hline {\textbf{Name}} & {\small{\textbf{ex}}} & {\small{\textbf{Genes}}} & {\textbf{Ratio}} & {\textbf{Size}} & {\small{\textbf{G-errs}}} & {\small{\textbf{B-errs}}} & {\textbf{Bound}} \\
\hline
{\small Colon} & 62 & 2000 &  0.42 & 1 & 12 &  11 & 33\\
{\small B\_MD} & 34 & 7129 &  0.10 & 1 &  7 &  7 & 20\\
{\small C\_MD} & 60 & 7129 & 0.08 & 5 & 21 & 20 & 45\\
{\small Leuk} & 72 & 7129 & 0.002 & 3 & 22 & 21 & 48\\
{\small Lung} & 52 & 918 & 0.12 & 1 & 3 & 3 & 18\\
{\small BreastER} & 49 & 7129 & 0.09 & 2 & 11 & 11 & 29\\
\hline
\end{tabular}
}
\caption{An illustration of the PAC-Bayes risk bound on a sample run of the PAC-Bayes algorithm.} \label{tab:results-PAC-Bayes-bound}
\end{center}
\end{table}

\subsection{A Note on the Risk Bound}\label{sec:note-on-risk-bound}

Note that the risk bounds are quite effective and their relevance
should not be misconstrued by observing the results in just the
current scenario. One of the most limiting factor in the current
analysis is the unavailability of microarray data with larger number
of examples. As the number of examples increase, the risk bound of
Theorem 5 gives tighter guarantees. Consider, for instance, if the datasets for the Lung and Colon Cancer had $500$ examples. A classifier with the same performance over 500 examples (i.e. with the same classification accuracy and number of features as currently) would
have a bound of about 12 and 30 percent error instead of current 34.6 and 54.6 percent respectively. This only illustrates how the bound can be more effective as a guarantee when used on datasets with more examples. Similarly, a dataset of 1000 examples for Breast Cancer with a similar performance can have a bound of about 30 percent instead of current 63 percent. Hence, the current limitation in the practical application of the bound comes from limited data availability. As the number of examples increase, the bounds provides tighter guarantees and become more significant.

\section{Analysis}\label{sec:analysis}

The results clearly show that even though
``Occam'' and ``SC'' are able to find sparse classifiers (with
very few genes), they are not able to obtain acceptable
classification accuracies. One possible explanation is that these
two approaches focus on the most succinct classifier with their
respective criterion. The Sample compression approach tries to
minimize the number of genes used but does not take into account the
magnitude of the separating margin and hence compromises accuracy.
On the other hand, the Occam's Razor approach tries to find a
classifier that depends on margin \emph{only indirectly}. Approaches based on sample compression as well as minimum description length have shown encouraging results in various domains. An alternate explanation for their suboptimal performance here can be seen in terms of extremely limited sample sizes. As a result, the gain in accuracy does not offset the cost of adding additional features in the conjunction. The PAC-Bayes approach seems to alleviate these problems by
performing a significant margin-sparsity tradeoff. That is, the advantage of adding a new feature is seen in terms of a combination of the gain in both margin and the empirical risk. This can be compared to the strategy used by the regularization approaches. The
classification accuracy of PAC-Bayes algorithm is competitive with
the best performing classifier but has an added advantage, quite
importantly, of using \emph{very few genes}.

For the PAC-Bayes approach, we expect the Bayes classifier to
generally perform better than the Gibbs classifier. This is
reflected to some extent in the empirical results for Colon, C\_MD
and Leukaemia datasets. However, there is no means to prove that
this will always be the case. It should be noted that there exist
several different utility functions that we can use for each of the
proposed learning approaches. We have tried some of these and
reported results only for the ones that were found to be the best
(and discussed in the description of the corresponding learning
algorithms).

A noteworthy observation with regard to Adaboost is that the gene
subset identified by this algorithm almost always include the
ones found by the proposed PAC-Bayes approach for decision stumps.
Most notably, the \emph{only} gene \emph{Cyclin D1}, a well known
marker for Cancer, found for the lung cancer dataset is the most
discriminating factor and is commonly found by both approaches. In
both cases, the size of the classifier is almost always restricted to $1$.
These observations not only give insights into the absolute peaks
worth investigating but also experimentally validates the proposed approaches.

Finally, many of the genes identified by the \emph{final}\footnote{This is the classifier learned after choosing
the best parameters using nested 5-fold CV and trained on the full dataset.} PAC-Bayes
classifier include some prominent markers for the corresponding
diseases as detailed below. 

\subsection{Biological Relevance of the Selected Features}\label{sec:bio-sig}

Table~\ref{tab:genes} details the genes identified by the \emph{final} PAC-Bayes
classifier learned over each dataset after the parameter selection
phase. There are some prominent markers identified by the
classifier. Some of the main genes identified by the PAC-Bayes
approach are the ones identified by previous studies for each
disease--- giving confidence in the proposed approach. Some of the
discovered genes in this case include \emph{Human monocyte-derived
neutrophil-activating protein (MONAP) mRNA} in the case of Colon
Cancer dataset and \emph{oestrogen receptor} in the case of Breast
Cancer data, \emph{D79205\_at-Ribosomal protein L39},
\emph{D83542\_at-Cadherin-15} and \emph{U29195\_at-NPTX2 Neuronal
pentraxin II} in the case of Medulloblastomas datasets B\_MD and
C\_MD. Other genes identified have biological relevance, for
instance, the identification of \emph{Adipsin}, \emph{LAF-4} and
\emph{HOX1C} with regard to ALL/AML by our algorithm is in agreement
with that of the findings
of~\citet{chow-et-al-01},~\citet{hiwatari-et-al-03}
and~\citet{Lawrence-92} respectively and the studies that followed.

\begin{table}[t]
\begin{center}
\resizebox{11cm}{!}{
\begin{tabular}{|l|l|}
\hline \textbf{Dataset} & \textbf{Gene(s) identified by PAC-Bayes Classifier}\\
\hline {\small Colon}    & 1. \small{ Hsa\.627 M26383-Human
monocyte-derived
neutrophil-activating protein (MONAP) mRNA}\\
{\small B\_MD}& 1. D79205\_at-Ribosomal protein L39\\
{\small C\_MD}    & 1. \small{ S71824\_at-Neural Cell Adhesion
Molecule, Phosphatidylinositol-Linked Isoform Precursor}\\
& 2. D83542\_at-Cadherin-15\\
& 3. U29195\_at-NPTX2 Neuronal pentraxin II\\
& 4. X73358\_s\_at-HAES-1 mRNA\\
& 5. \small{L36069\_at-High conductance inward rectifier potassium channel alpha subunit mRNA}\\
{\small Leuk}   & 1. M84526\_at-DF D component of complement (adipsin)\\
& 2. U34360\_at-Lymphoid nuclear protein (LAF-4) mRNA\\
& 3. M16937\_at-Homeo box c1 protein, mRNA \\
{\small Lung} & 1. \small{GENE221X-IMAGE\_841641-cyclin D1 (PRAD1-parathyroid adenomatosis 1) Hs\.82932 AA487486} \\
{\small BreastER} & 1. X03635\_at,X03635- class C, 20 probes, 20 in all\_X03635 5885 - 6402\\
& Human mRNA for oestrogen receptor\\
& 2. L42611\_f\_at, L42611- class A, 20 probes, 20 in L42611 1374-1954,\\
& Homo sapiens keratin 6 isoform K6e \(KRT6E\) mRNA, complete cds\\
\hline
\end{tabular}
}
\end{center}
\caption{Genes Identified by the \emph{Final} PAC-Bayes Classifier}\label{tab:genes}
\end{table}

Further, in the case of breast cancer, Estrogen receptors (ER) have
shown to interact with BRCA1 to regulate VEGF transcription and
secretion in breast cancer cells~\citep{kawai-et-al-02}. These
interactions are further investigated by~\citet{ma-et-al-05}.
Further studies for ER have also been done. For
instance,~\citet{moggs-et-al-05} discovered 3 putative
estrogen-response elements in Keratin6 (the second gene identified
by the PAC-Bayes classifier in the case of BreastER data) in the
context of E2-responsive genes identified by microarray analysis of
MDA-MD-231 cells that re-express ER$_\alpha$. An important role
played by cytokeratins in cancer development is also widely known
(see for instance~\citet{gusterson-et-al-05}).

Furthermore, the importance of \emph{MONAP} in the case of colon
cancer and \emph{Adipsin} in the case of leukaemia data has further
been confirmed by various rank based algorithms as detailed
by~\citet{su-et-al-03} in the implementation of ``RankGene'', a
program that analyzes and ranks genes for the gene expression data
using eight ranking criteria including Information Gain (IG), Gini
Index (GI), Max Minority (MM), Sum Minority (SM), Twoing Rule (TR),
t-statistic (TT), Sum of variances (SV) and one-dimensional Support
Vector Machine (1S). In the case of Colon Cancer data, \emph{MONAP}
is identified as the top ranked gene by four of the eight criteria
(IG, SV, TR, GI), second by one (SM), eighth by one (MM) and in top
50 by 1S. Similarly, in the case of Leukaemia data, \emph{Adipsin}
is top ranked by 1S, fifth by SM, seventh by IG, SV, TR, GI and MM
and is in top 50 by TT. These observations provides a strong
validation for our approaches.

\emph{Cyclin} as identified in the case of Lung Cancer dataset is a
well known marker for cell division whose perturbations are
considered to be one of the major factors causing
cancer~\citep{driscoll-et-al-99,masaki-et-al-03}.

Finally, the discovered genes in the case of Medulloblastomas are
important with regard to the neuronal functioning (esp. S71824,
U29195 and L36039) and can have relevance for nervous system related
tumors.

\section{Conclusion}\label{sec:conclusion}

Learning from high-dimensional data such as that from DNA microarrays can be quite
challenging especially when the aim is to identify only a few attributes that characterizes the differences between two classes of data. We investigated the premise of learning conjunctions of \emph{decision stumps} and proposed three formulations based on
different learning principles. We observed that the approaches that aim solely to optimize sparsity or the message code with regard to the classifier's empirical risk limits the algorithm in terms of its generalization performance, at least in the present case of small dataset sizes. By trading-off the sparsity of the classifier with the separating margin in addition to the empirical risk, the PAC-Bayes approach seem to alleviate this problem to a significant extent. This allows the PAC-Bayes algorithm to yield competitive classification performance while at the same time utilizing significantly fewer attributes.

As opposed to the traditional feature selection methods, the proposed
approaches are accompanied by a \emph{theoretical justification of the
performance}. Moreover, the proposed algorithms \emph{embed the feature
selection as a part of the learning process} itself.\footnote{Note that
\citet{huang-chang-07} proposed one such approach. However, they
need multiple SVM learning runs. Hence, their method basically works as a wrapper.} Furthermore, the generalization error bounds are practical and can potentially guide the model (parameter) selection. When applied to classify DNA microarray data, the genes
identified by the proposed approaches are found to be biologically significant as experimentally validated by various studies, an empirical justification that the approaches can successfully perform meaningful feature selection. Consequently, this represents a significant improvement in the direction of successful integration of machine
learning approaches for use in high-throughput data to provide
\emph{meaningful, theoretically justifiable}, and \emph{reliable}
results. Such approaches that yield a compressed view in terms of a
small number of biological markers can lead to a targeted and well
focussed study of the issue of interest. For instance, the approach can be utilized in identifying gene subsets from the microarray experiments that should be further validated using focused RT-PCR techniques which are otherwise both costly and impractical to perform on the full set of genes.

Finally, as mentioned previously, the approaches presented in this wor have a wider relevance, and can have significant implications in the direction of designing theoretically justified feature selection algorithms. These are one of the few approaches that combines the feature selection with the learning process \emph{and} provide generalization guarantees over the resulting classifiers \emph{simultaneously}. This property assumes even more significance in the wake of limited size of microarray datasets since it limits the amount of empirical evaluation that can be reliably performed otherwise. Most natural extensions of the approaches and the learning bias proposed here would be in other similar domains including other forms of microarray experiments such as Chromatin Immunoprecipitation promoter arrays (chIP-Chip) and from Protein arrays. Within the same learning settings, other learning biases can also be explored such as classifiers represented by features or sets of features built on subsets of attributes.


%

\appendices


\ifCLASSOPTIONcompsoc
  \section*{Acknowledgments}
\else
  \section*{Acknowledgment}
\fi
This work was supported by the National Science and Engineering Research Council (NSERC)
of Canada [Discovery Grant No. 122405 to MM], the Canadian Institutes of Health Research
[operating grant to JC, training grant to MS while at CHUL] and the Canada Research Chair in
Medical Genomics to JC.

\ifCLASSOPTIONcaptionsoff
  \newpage
\fi



\bibliographystyle{plainnat}
\bibliography{mohak-bib}

\begin{thebibliography}{38}
\providecommand{\natexlab}[1]{#1}
\providecommand{\url}[1]{\texttt{#1}}
\expandafter\ifx\csname urlstyle\endcsname\relax
  \providecommand{\doi}[1]{doi: #1}\else
  \providecommand{\doi}{doi: \begingroup \urlstyle{rm}\Url}\fi

\bibitem[Alon et~al.(1999)Alon, Barkai, Notterman, Gish, Ybarra, Mack, and
  Levine]{a-99}
U.~Alon, N.~Barkai, D.A. Notterman, K.~Gish, S.~Ybarra, D.~Mack, and A.J.
  Levine.
\newblock Broad patterns of gene expression revealed by clustering analysis of
  tumor and normal colon tissues probed by oligonucleotide arrays.
\newblock \emph{Proc. Natl. Acad. Sci. USA}, 96\penalty0 (12):\penalty0
  6745--6750, 1999.

\bibitem[Ambroise and McLachlan(2002)]{a-02}
C.~Ambroise and G.~J. McLachlan.
\newblock Selection bias in gene extraction on the basis of microarray
  gene-expression data.
\newblock \emph{Proc. Natl. Acad. Sci. USA}, 99\penalty0 (10):\penalty0
  6562--6566, 2002.

\bibitem[Blum and Langford(2003)]{bl-03}
Avrim Blum and John Langford.
\newblock {PAC-MDL} bounds.
\newblock In \emph{Proceedings of 16th Annual Conference on Learning Theory,
  {COLT} 2003, {\rm Washington, DC, August 2003}}, volume 2777 of \emph{Lecture
  Notes in Artificial Intelligence}, pages 344--357. Springer, Berlin, 2003.

\bibitem[Blumer et~al.(1987)Blumer, Ehrenfeucht, Haussler, and
  Warmuth]{behw-87}
A.~Blumer, A.~Ehrenfeucht, D.~Haussler, and M.~Warmuth.
\newblock Occam's razor.
\newblock \emph{Information Processing Letters}, 24:\penalty0 377--380, 1987.

\bibitem[Chow et~al.(2001)Chow, Moler, and Mian]{chow-et-al-01}
M.~L. Chow, E.~J. Moler, and I.~S. Mian.
\newblock Identifying marker genes in transcription profiling data using a
  mixture of feature relevance experts.
\newblock \emph{Phsiol Genomics}, 5\penalty0 (2):\penalty0 99--111, 2001.

\bibitem[Driscoll et~al.(1999)Driscoll, Buckley, Barsky, Weinberg, Anderson,
  and Warburton]{driscoll-et-al-99}
B.~Driscoll, S.~Buckley, L.~Barsky, K.~Weinberg, K.~D. Anderson, and
  D.~Warburton.
\newblock Abrogation of cyclin {D}1 expression predisposes lung cancer cells to
  serum deprivation-induced apoptosis.
\newblock \emph{Am J Phsiol}, 276\penalty0 (4 Pt 1):\penalty0 L679--687, 1999.

\bibitem[Eisen and Brown(1999)]{eisen-99}
M.~Eisen and P.~Brown.
\newblock {DNA} arrays for analysis of gene expression.
\newblock \emph{Methods Enzymology}, 303:\penalty0 179--205, 1999.

\bibitem[Furey et~al.(2000)Furey, Cristianini, Duffy, Bednarski, Schummer, and
  Haussler]{fcdbsh-00}
T.~S. Furey, N.~Cristianini, N.~Duffy, D.~W. Bednarski, M.~Schummer, and
  D.~Haussler.
\newblock Support vector machine classification and validation of cancer tissue
  samples using microarray expression data.
\newblock \emph{Bioinformatics}, 16:\penalty0 906--914, 2000.

\bibitem[Garber et~al.(2001)Garber, Troyanskaya, Schluens, Petersen, Thaesler,
  Pacyna-Gengelbach, van~de Rijn, Rosen, Perou, Whyte, Altman, Brown, Botstein,
  and Petersen]{g-01}
M.~E. Garber, O.~G. Troyanskaya, K.~Schluens, S.~Petersen, Z.~Thaesler,
  M.~Pacyna-Gengelbach, M.~van~de Rijn, G.~D. Rosen, C.~M. Perou, R.~I. Whyte,
  R.~B. Altman, P.~O. Brown, D.~Botstein, and I.~Petersen.
\newblock Diversity of gene expression in adenocarcinoma of the lung.
\newblock \emph{Proc. Natl. Acad. Sci. USA}, 98\penalty0 (24):\penalty0
  13784--13789, 2001.

\bibitem[Golub et~al.(1999)Golub, Slonim, Tamayo, Huard, Gaasenbeek, Mesirov,
  Coller, Loh, Downing, Caligiuri, Bloomfield, and Lander]{g-99}
T.~R. Golub, D.~K. Slonim, P.~Tamayo, C.~Huard, M.~Gaasenbeek, J.~P. Mesirov,
  H.~Coller, M.~L. Loh, J.~R. Downing, M.~A. Caligiuri, C.~D. Bloomfield, and
  E.~S. Lander.
\newblock Molecular classification of cancer: class discovery and class
  prediction by gene expression monitoring.
\newblock \emph{Science}, 286\penalty0 (5439):\penalty0 531--537, 1999.

\bibitem[Gusterson et~al.(2005)Gusterson, Ross, Heath, and
  Stein]{gusterson-et-al-05}
B.~A. Gusterson, D.~T. Ross, V.~J. Heath, and T.~Stein.
\newblock Basal cytokeratins and their relationship to the cellular origin and
  functional classification of breast cancer.
\newblock \emph{Breast Cancer Research}, 7:\penalty0 143--148, 2005.

\bibitem[Guyon et~al.(2002)Guyon, Weston, Barnhill, and Vapnik]{gwbv-02}
Isabelle Guyon, Jason Weston, Stephen Barnhill, and Vladimir Vapnik.
\newblock Gene selection for cancer classification using support vector
  machines.
\newblock \emph{Machine Learning}, 46:\penalty0 389--422, 2002.

\bibitem[Haussler(1988)]{h-88}
D.~Haussler.
\newblock Quantifying inductive bias: {AI} learning algorithms and {V}aliant's
  learning framework.
\newblock \emph{Artificial Intelligence}, 36:\penalty0 177--221, 1988.

\bibitem[Hiwatari et~al.(2003)Hiwatari, Taki, Taketani, Taniwaki, Sugita,
  Okuya, Eguchi, Ida, and Hayashi]{hiwatari-et-al-03}
Mitsuteru Hiwatari, Tomohiko Taki, Takeshi Taketani, Masafumi Taniwaki, Kenichi
  Sugita, Mayuko Okuya, Mitsuoki Eguchi, Kohmei Ida, and Yasuhide Hayashi.
\newblock Fusion of an {AF}4-related gene, {LAF}4, to {MLL} in childhood acute
  lymphoblastic leukemia with t(2;11)(q11;q23).
\newblock \emph{Oncogene}, 22\penalty0 (18):\penalty0 2851--2855, 2003.

\bibitem[Huang and Chang(2007)]{huang-chang-07}
H.~Huang and F.~Chang.
\newblock {ESVM}: Evolutionary support vector machine for automatic feature
  selection and classification of microarray data.
\newblock \emph{Biosystems}, 90\penalty0 (2):\penalty0 516--528, 2007.

\bibitem[Kawai et~al.(2002)Kawai, Li, Chun, S, and Avraham]{kawai-et-al-02}
H.~Kawai, H.~Li, P.~Chun, S.~Avraham S, and H.~K. Avraham.
\newblock Direct interaction between {BRCA}1 and the estrogen receptor
  regulates vascular endothelial growth factor ({VEGF}) transcription and
  secretion in breast cancer cells.
\newblock \emph{Oncogene}, 21\penalty0 (50):\penalty0 7730--7739, 2002.

\bibitem[Kuzmin and Warmuth(2007)]{kw-07}
Dima Kuzmin and Manfred~K. Warmuth.
\newblock Unlabeled compression schemes for maximum classes.
\newblock \emph{J. Mach. Learn. Res.}, 8:\penalty0 2047--2081, 2007.
\newblock ISSN 1533-7928.

\bibitem[Langford(2005)]{l-05}
John Langford.
\newblock Tutorial on practical prediction theory for classification.
\newblock \emph{Journal of Machine Learning Research}, 3:\penalty0 273--306,
  2005.

\bibitem[Lawrence and Largman(1992)]{Lawrence-92}
H.~J. Lawrence and C.~Largman.
\newblock Homeobox genes in normal hematopoiesis and leukemia.
\newblock \emph{Blood}, 80\penalty0 (10):\penalty0 2445--2453, 1992.

\bibitem[Lipshutz et~al.(1999)Lipshutz, Fodor, Gingeras, and
  Lockhart]{Lipshutz-et-al-99}
R.~Lipshutz, S.~Fodor, T.~Gingeras, and D.~Lockhart.
\newblock High density synthetic oligonucleotide arrays.
\newblock \emph{Nature Genetics}, 21\penalty0 (1 Suppl):\penalty0 20--24, 1999.

\bibitem[Ma et~al.(2005)Ma, Tomita, Fan, Wu, Tong, Zhao, Song, Goldberg, and
  Rosen]{ma-et-al-05}
Y.~X. Ma, Y.~Tomita, S.~Fan, K.~Wu, Y.~Tong, Z.~Zhao, L.~N. Song, I.~D.
  Goldberg, and E.~M. Rosen.
\newblock Structural determinants of the {BRCA}1 : estrogen receptor
  interaction.
\newblock \emph{Oncogene}, 24\penalty0 (11):\penalty0 1831--1846, 2005.

\bibitem[{Marchand} and {Shah}(2005)]{ms-05}
Mario {Marchand} and Mohak {Shah}.
\newblock {PAC}-bayes learning of conjunctions and classification of
  gene-expression data.
\newblock In Lawrence~K. Saul, Yair Weiss, and {L\'{e}on} Bottou, editors,
  \emph{Advances in Neural Information Processing Systems 17}, pages 881--888.
  MIT Press, Cambridge, MA, 2005.

\bibitem[Marchand and Shawe-Taylor(2002)]{ms-02}
Mario Marchand and John Shawe-Taylor.
\newblock The set covering machine.
\newblock \emph{Journal of Machine Learning Reasearch}, 3:\penalty0 723--746,
  2002.

\bibitem[Marchand and Sokolova(2005)]{msok-05}
Mario Marchand and Marina Sokolova.
\newblock Learning with decision lists of data-dependent features.
\newblock \emph{Journal of Machine Learning Reasearch}, 6:\penalty0 427--451,
  2005.

\bibitem[Masaki et~al.(2003)Masaki, Shiratori, Rengifo, Igarashi, Yamagata,
  Kurokohchi, Uchida, Miyauchi, Yoshiji, Watanabe, Omata, and
  Kuriyama]{masaki-et-al-03}
T.~Masaki, Y.~Shiratori, W.~Rengifo, K.~Igarashi, M.~Yamagata, K.~Kurokohchi,
  N.~Uchida, Y.~Miyauchi, H.~Yoshiji, S.~Watanabe, M.~Omata, and S.~Kuriyama.
\newblock Cyclins and cyclin-dependent kinases: Comparative study of
  hepatocellular carcinoma versus cirrhosis.
\newblock \emph{Hepatology}, 37\penalty0 (3):\penalty0 534--543, 2003.

\bibitem[McAllester(2003)]{m-03}
David McAllester.
\newblock {PAC}-{B}ayesian stochastic model selection.
\newblock \emph{Machine Learning}, 51:\penalty0 5--21, 2003.
\newblock A priliminary version appeared in proceedings of COLT'99.

\bibitem[McAllester(1999)]{m-99}
David McAllester.
\newblock Some {PAC}-{B}ayesian theorems.
\newblock \emph{Machine Learning}, 37:\penalty0 355--363, 1999.

\bibitem[Moggs et~al.(2005)Moggs, Murphy, Lim, Moore, Stuckey, Antrobus,
  Kimber, and Orphanides]{moggs-et-al-05}
J.~G. Moggs, T.~C. Murphy, F.~L. Lim, D.~J. Moore, R.~Stuckey, K.~Antrobus,
  I.~Kimber, and G.~Orphanides.
\newblock Anti-proliferative effect of estrogen in breast cancer cells that
  re-express {ER}alpha is mediated by aberrant regulation of cell cycle genes.
\newblock \emph{Journal of Molecular Endocrinology}, 34:\penalty0 535--551,
  2005.

\bibitem[Pomeroy et~al.(2002)Pomeroy, Tamayo, Gaasenbeek, Sturla, Angelo,
  McLaughlin, Kim, Goumnerova, Black, Lau, Allen, Zagzag, Olson, Curran,
  Wetmore, Biegel, Poggio, Mukherjee, Rifkin, Califano, Stolovitzky, Louis,
  Mesirov, Lander, and Golub]{p-02}
S.~L. Pomeroy, P.~Tamayo, M.~Gaasenbeek, L.~M. Sturla, M.~Angelo, M.~E.
  McLaughlin, J.~Y. Kim, L.~C. Goumnerova, P.~M. Black, C.~Lau, J.~C. Allen,
  D.~Zagzag, J.~M. Olson, T.~Curran, C.~Wetmore, J.~A. Biegel, T.~Poggio,
  S.~Mukherjee, R.~Rifkin, A.~Califano, G.~Stolovitzky, D.~N. Louis, J.~P.
  Mesirov, E.~S. Lander, and T.~R. Golub.
\newblock Prediction of central nervous system embryonal tumour outcome based
  on gene expression.
\newblock \emph{Nature}, 415\penalty0 (6870):\penalty0 436--442, 2002.

\bibitem[Seeger(2002)]{s-02}
Matthias Seeger.
\newblock {PAC}-{B}ayesian generalization bounds for gaussian processes.
\newblock \emph{Journal of Machine Learning Research}, 3:\penalty0 233--269,
  2002.

\bibitem[Shah and Corbeil(2010)]{sc-09}
Mohak Shah and Jacques Corbeil.
\newblock A general framework for analyzing data from two short time-series
  microarray experiments.
\newblock \emph{IEEE/ACM Transactions on Computational Biology and
  Bioinformatics}, to appear, 2010.
\newblock \doi{http://doi.ieeecomputersociety.org/10.1109/TCBB.2009.51}.

\bibitem[Song et~al.(2007)Song, Bedo, Borgwardt, Gretton, and Smola]{song-07}
L.~Song, J.~Bedo, K.~M. Borgwardt, A.~Gretton, and A.~Smola.
\newblock Gene selection via the {BAHSIC} family of algorithms.
\newblock \emph{Bioinformatics}, 23\penalty0 (13):\penalty0 490--498, 2007.

\bibitem[Su et~al.(2003)Su, Murali, Pavlovic, Schaffer, and Kasif]{su-et-al-03}
Yang Su, T.M. Murali, Vladimir Pavlovic, Michael Schaffer, and Simon Kasif.
\newblock Rank{G}ene: identification of diagnostic genes based on expression
  data.
\newblock \emph{Bioinformatics}, 19\penalty0 (12):\penalty0 1578--1579, 2003.

\bibitem[Tibshirani et~al.(2003)Tibshirani, Hastie, Narasimhan, and
  Chu]{Tibshirani-et-al-03}
R.~Tibshirani, T.~Hastie, B.~Narasimhan, and G.~Chu.
\newblock Class predicition by nearest shrunken centroids with applications to
  dna microarrays.
\newblock \emph{Statistical Science}, 18:\penalty0 104--117, 2003.

\bibitem[Wang et~al.(2007)Wang, Chu, and Xie]{wang-et-al-07}
Lipo Wang, Feng Chu, and Wei Xie.
\newblock Accurate cancer classification using expressions of very few genes.
\newblock \emph{IEEE/ACM Trans. Comput. Biol. Bioinformatics}, 4\penalty0
  (1):\penalty0 40--53, 2007.
\newblock ISSN 1545-5963.
\newblock \doi{http://dx.doi.org/10.1109/TCBB.2007.1006}.

\bibitem[West et~al.(2001)West, Blanchette, Dressman, Huang, Ishida, Spang,
  Zuzan, Jr, Marks, and Nevins]{w-01}
M.~West, C.~Blanchette, H.~Dressman, E.~Huang, S.~Ishida, R.~Spang, H.~Zuzan,
  J.~A.~Olson Jr, J.~R. Marks, and J.~R. Nevins.
\newblock Predicting the clinical status of human breast cancer by using gene
  expression profiles.
\newblock \emph{Proc. Natl. Acad. Sci. USA}, 98\penalty0 (20):\penalty0
  11462--11467, 2001.

\bibitem[Witten and Frank(2005)]{wf-05}
Ian~H. Witten and Eibe Frank.
\newblock \emph{Data Mining: Practical machine learning tools and techniques,
  2nd Ed.}
\newblock Morgan Kaufmann, San Francisco, 2005.

\bibitem[Zhang and Yu(2005)]{zhang-yu-05}
T.~Zhang and B.~Yu.
\newblock Boosting with early stopping: Convergence and consistency.
\newblock \emph{The Annals of Statistics}, 33:\penalty0 1538--1579, 2005.

\end{thebibliography}
\end{document}